\documentclass[conference]{IEEEtran}
\usepackage[utf8]{inputenc}
\usepackage{graphicx} 
\usepackage{amsmath}
\usepackage{array}
\usepackage{algorithm}
\usepackage{algorithmic}

\makeatletter

\def\ps@IEEEtitlepagestyle{
  \def\@oddfoot{\mycopyrightnotice}
  \def\@evenfoot{}
}
\def\mycopyrightnotice{
  {\footnotesize 979-8-3315-3197-3/24/\$31.00 ~\copyright~2024 IEEE\hfill} 
  \gdef\mycopyrightnotice{}
}

\@ifundefined{showcaptionsetup}{}{
 \PassOptionsToPackage{caption=false}{subfig}}
\usepackage{subfig}
\makeatother

\usepackage{eso-pic}
\newcommand\AtPageUpperMyright[1]{\AtPageUpperLeft{
 \put(\LenToUnit{0.5\paperwidth},\LenToUnit{-1cm}){
     \parbox{0.5\textwidth}{\raggedleft\fontsize{9}{11}\selectfont #1}}
 }}
\newcommand{\conf}[1]{
\AddToShipoutPictureBG*{
\AtPageUpperMyright{#1}
}
}

\begin{document}

\title{AquaAugmentor: A Novel Feature Augmentation Algorithm for Water Potability Prediction}
\conf{2024 6\textsuperscript{th} International Conference on Sustainable Technologies for Industry 5.0 (STI), 14-15 December, Dhaka} 


\author{\IEEEauthorblockN{
Muntasir Tabasum\textsuperscript{1}, 
Al Zadid Sultan Bin Habib\textsuperscript{2},
Tanpia Tasnim\textsuperscript{3},
Md. Ekramul Islam\textsuperscript{4}, \\
Md Younus Ahamed\textsuperscript{5},
Md Asif Bin Syed\textsuperscript{6}, 
}
  \IEEEauthorblockA{
    \textsuperscript{1,2,5,6}West Virginia University, Morgantown, WV 26506, USA\\
    \textsuperscript{3}Green University of Bangladesh, Narayanganj-1461, Dhaka, Bangladesh\\
    \textsuperscript{4}Stamford University Bangladesh, Dhaka-1217, Bangladesh\\
    Email: \{mt00079\textsuperscript{1}, ah00069\textsuperscript{2}, ma00087\textsuperscript{5}, ms00110\textsuperscript{6}\}@mix.wvu.edu,
    tanpia@cse.green.edu.bd\textsuperscript{3},
    eislam706@gmail.com\textsuperscript{4}
}
}

\maketitle

\begin{abstract}
Access to potable water is crucial for health, economic development, and sustainability. However, accurately classifying water quality remains a significant challenge due to the complexity and variability of water source data. This paper addresses the challenge of predicting water potability through machine learning and deep learning algorithms. It introduces a novel feature augmentation algorithm, AquaAugmentor, to enhance the predictive performance of these models for low-dimensional datasets. Utilizing a dataset that includes chemical attributes of water, such as pH, hardness, solids, chloramines, sulfate, and others. This study evaluates the performance of the models with and without AquaAugmentor. Each model applied to classify water as potable or non-potable and its performance is then evaluated and compared based on test accuracy and AUC score. The results highlight the strengths and limitations of our proposed algorithm, providing insights into the most effective techniques for improving the predictive performance of water quality classification. This study contributes to the broader efforts of ensuring safe water access and serves as a framework for employing machine learning in environmental quality assessments. The findings aim to assist researchers, policymakers, and public health officials in making informed decisions based on reliable machine learning predictions.
\end{abstract}


\section{Introduction}
Access to clean and safe drinking water is a basic human right, yet it remains a major challenge around the world. The World Health Organization (WHO) reports that more than two billion people live in water-stressed countries, where access to safe water is either limited or entirely unavailable \cite{b1}. Efficient testing and verification of water quality is crucial for both public health and environmental sustainability. However, traditional water quality testing methods can be time-consuming, often requiring specialized equipment and trained personnel. As a result, there is increasing interest in using advanced technologies to streamline and improve the accuracy of water quality testing. Innovative solutions, such as real-time monitoring systems, biosensors, and machine learning algorithms, are being explored to provide rapid and precise assessments of water potability. These advancements have the potential to greatly reduce the risks associated with contaminated water, ensuring safer drinking water for more people and supporting global efforts to achieve the United Nations’ Sustainable Development Goals (SDGs) related to clean water and sanitation \cite{b19, b20}.\newline
Machine learning offers promising solutions to these challenges by enabling the rapid, cost-effective, and accurate classification of water potability based on chemical and physical parameters. Recent studies have demonstrated the potential of various machine learning algorithms in environmental monitoring and public health applications, including water quality assessment \cite{b2}, \cite{b3}. For instance, Support Vector Machines (SVM) and neural networks have been particularly noted for their predictive accuracy in complex environmental data scenarios \cite{b2}. These models can process large datasets to identify patterns and correlations that might be missed by traditional methods, providing more reliable predictions about water quality. Additionally, machine learning models can be continuously updated and improved with new data, ensuring they remain effective as environmental conditions and pollution sources change. Integrating machine learning with sensor networks and real-time data collection can develop more responsive and adaptive water quality monitoring systems. These advancements hold great potential for enhancing public health protection, reducing the incidence of waterborne diseases, and supporting the sustainable management of water resources \cite{b21}.\newline
This research explores and compares the effectiveness of various machine learning and deep learning models in classifying water as potable or non-potable. Each model’s performance was evaluated using test accuracy and AUC score to identify the most suitable approaches for real-world applications. Our novel contribution, AquaAugmentor, a feature augmentation algorithm, significantly enhances the predictive performance of these models, especially for low-dimensional datasets. This advancement highlights the potential of integrating advanced computational techniques to improve water quality assessment, contributing to broader environmental science efforts to ensure safe water access. In this paper, section \ref{sec:2} highlights the literature review of related existing works. Section \ref{sec:3} elaborates the methodological framework. Section \ref{sec:4} discusses the results, and section \ref{sec:5} concludes the paper with a blueprint of future works. 
\section{Related Work}
\label{sec:2}
Access to potable water is a global concern that impacts health, sustainability, and economic development. Recent studies have focused on evaluating and classifying water potability using machine learning or deep learning, exploring various models to predict water quality from chemical and physical parameters.\newline
Mukati et al. explored the effectiveness of ensemble methods like Random Forest, Decision Trees, and XGBoost, emphasizing the improved predictive accuracy these methods provide over traditional statistical techniques \cite{b4}. This aligns with findings from De Luna et al., who tested AdaBoost, XGBoost, and ExtraTree Classifier, finding that more complex ensemble models could sometimes significantly enhance prediction accuracy, particularly in handling non-linear and complex dataset structures \cite{b5}. Gao et al. employed binomial Logistic Regression and K-Nearest Neighbor (KNN) algorithms, demonstrating their suitability for smaller datasets and emphasizing the importance of each water quality feature’s independent influence on potability \cite{b6}. SVM was particularly noted for its high accuracy in datasets with clear margin separations. Khanna et al. introduced a novel approach using deep learning for water potability classification, highlighting its capability to capture deeper insights from complex interdependencies among water quality parameters \cite{b9}. Their findings suggest that deep learning could offer a significant step forward in predictive accuracy and reliability. Yusuf et al. demonstrated the effectiveness of Random Forest and Decision Trees in achieving higher classification accuracy, underscoring the significance of feature selection in model performance \cite{b10}. Deep learning models, especially Artificial Neural Networks (ANNs) and Long Short-Term Memory (LSTM) networks, have been applied to predict water quality with high precision, addressing complex non-linear relationships within the data. Such models were highlighted by Suleiman et al. for their exceptional ability to classify groundwater potability, reflecting their growing application in environmental sciences \cite{b11}. As Alipio discussed, integrating Internet of Things (IoT) technologies with machine learning models transforms water quality monitoring. This approach leverages real-time data acquisition and machine learning-driven analytics to enhance the responsiveness and accuracy of water potability assessments \cite{b12}. Comparative studies of machine learning algorithms reveal varying strengths across different techniques. Haq et al. compared several machine learning algorithms and found that models like XGBoost and ExtraTree classifiers provided superior performance due to their robust handling of diverse datasets \cite{b13}. Argreen et al. \cite{b14} explore the application of deep learning models for water potability classification in rural areas of the Philippines, highlighting the effectiveness of these models in enhancing water quality monitoring. Ahmad Musleh \cite{b15} conducted a comprehensive comparative study of six machine learning algorithms for water potability classification, finding that Random Forest and J48 achieved the highest accuracy and precision in predicting water quality. Alipio \cite{b16} developed a data-driven IoT-based system for real-time water quality monitoring and potability classification for rural areas. He matched his results with conventional laboratory tests and demonstrated high accuracy and minimal data transmission delays. Patel et al. \cite{b17} proposed a machine learning-based water potability prediction model using the Synthetic Minority Over-sampling Technique (SMOTE) and explainable AI techniques, achieving significant accuracy improvements and providing insights into the feature importance of water quality assessment. Abuzir and Abuzir \cite{b18} demonstrated that Multi-Layer Perceptron (MLP) outperforms J48 and Naïve Bayes in water quality classification, highlighting the significance of feature selection and dimensionality reduction using PCA for improving prediction accuracy.\newline
These existing works incorporate a broad spectrum of current research, highlighting the dynamic nature of machine learning applications in environmental monitoring. Integrating advanced computational techniques, such as deep learning and IoT-enabled machine learning models, is noteworthy and offers promising avenues for enhancing water quality assessment. However, these approaches are often dataset-specific and vary significantly across different datasets. The dimensionality of data can differ, posing challenges for achieving good accuracy with low-dimensional datasets. Our proposed AquaAugmentor algorithm addresses this challenge by leveraging the concept of feature augmentation, providing a robust solution for improving predictive performance with low-dimensional datasets. Our work advances sustainable water management with AI-driven predictions, crucial for industry 5.0.
\section{Methodology}
\label{sec:3}
\subsection{Input Dataset}
The dataset \cite{b22} used in this study comprises 3,276 samples, each with nine chemical attributes of water: pH, Hardness, Solids, Chloramines, Sulfate, Conductivity, Organic Carbon, Trihalomethanes, and Turbidity. These features are crucial for determining water potability. The dataset includes potable and non-potable classes, with 1,998 samples labeled as non-potable and 1,278 samples labeled as potable. Before proceeding to preprocessing and analysis, we performed mean imputation to address missing values, ensuring a complete and reliable dataset for accurate modeling.
\subsection{Feature Augmentation}
\begin{figure}
\centering
\includegraphics[width=0.8\linewidth, height=0.9\linewidth]{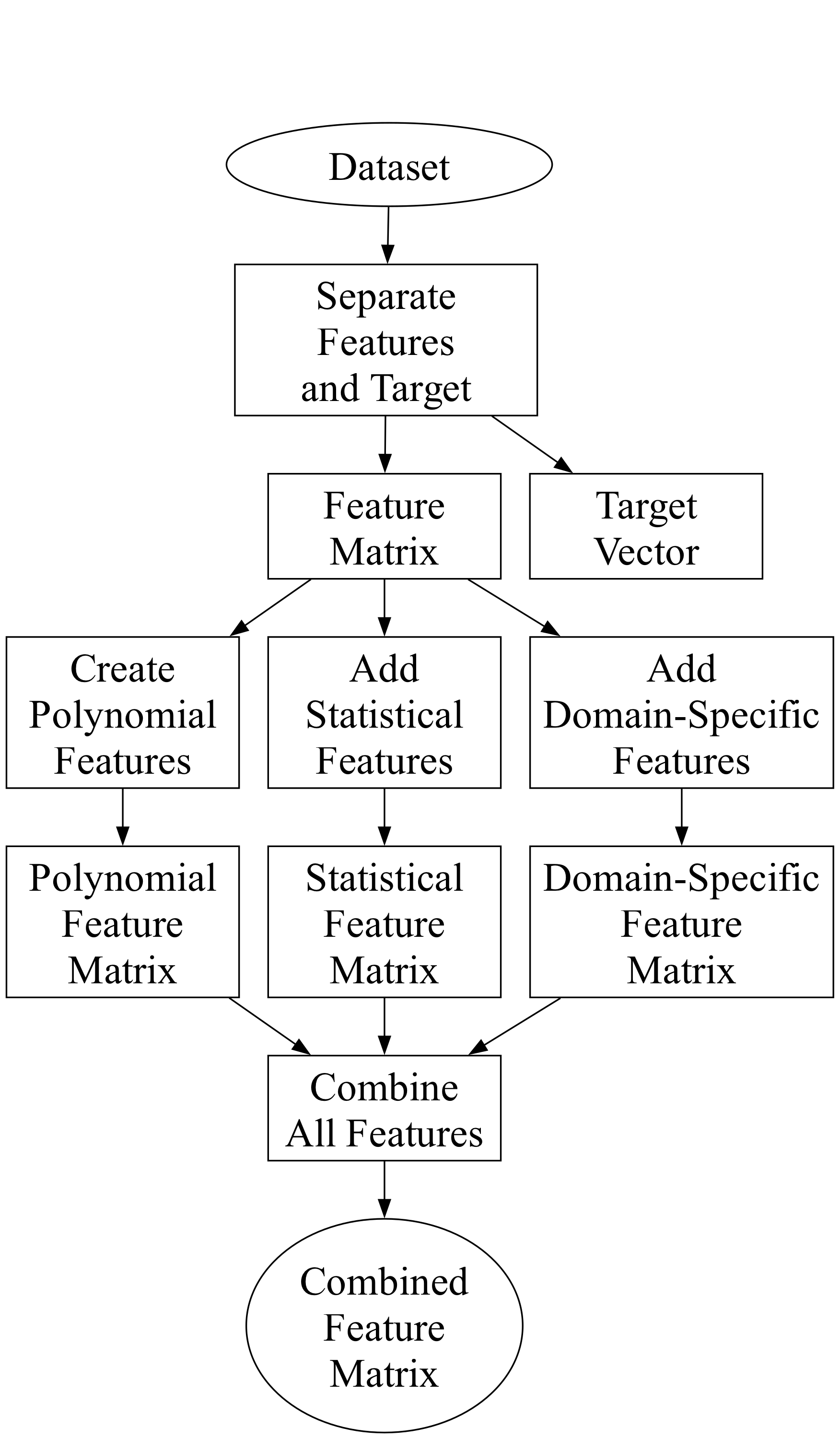}
\caption{Workflow of the AquaAugmentor algorithm.}
\label{fig}
\end{figure}

Feature augmentation is the process of enhancing the original feature set by generating new features through various methods, including polynomial combinations, statistical summarizations, and domain-specific calculations.\\
\textbf{Polynomial Features:} It generates new features by taking polynomial combinations of existing features up to a specified degree.\\
\textbf{Statistical Features:} It adds features that summarize the data, such as mean, standard deviation, minimum, and maximum values for each sample.\\
\textbf{Domain Specific Features:} It creates new features based on domain knowledge, such as ratios and other relationships between existing features.\\
By combining these methods, feature augmentation significantly expands the feature space, potentially improving the model's ability to capture complex patterns in the data. In this study, we applied feature augmentation to enhance the original dataset. This involved generating polynomial features, adding statistical summaries, and creating domain-specific ratios. This comprehensive approach aimed to provide the model with a richer and more informative feature set, thereby improving its performance. Given a dataset \( D \) with \( n \) samples and \( m \) features, where the last column is the target variable \( y \):
\begin{equation}
\mathbf{X} = D[:, :m]
\label{eq:X}
\end{equation}
\begin{equation}
\mathbf{y} = D[:, m]
\label{eq:y}
\end{equation}
Where, \( \mathbf{X} \) is the feature matrix of shape \( (n, m) \) and \( \mathbf{y} \) is the target vector of shape \( (n,) \).
Given a feature matrix \( \mathbf{X} \) of shape \( (n, m) \), the polynomial feature matrix \( \mathbf{X}_{\text{poly}} \) includes all polynomial combinations of the features up to a specified degree \( d \):
\begin{equation}
\label{eq:poly}
\mathbf{X}_{\text{poly}} = \text{Poly}(\mathbf{X}, d)    
\end{equation}
Where \( \text{Poly}(\mathbf{X}, d) \) generates a new matrix including all polynomial combinations of features in \( \mathbf{X} \) up to degree \( d \) and the shape of \( \mathbf{X}_{\text{poly}} \) depends on the number of polynomial features generated. For each sample \( i \) in the feature matrix \( \mathbf{X} \), calculate the mean (\( \mu_i \)), standard deviation (\( \sigma_i \)), minimum (\( \min_i \)), and maximum (\( \max_i \)):
\begin{equation}
\label{eq:mean}
\mu_i = \frac{1}{m} \sum_{j=1}^{m} X_{ij}
\end{equation}
\begin{equation}
\label{eq:std}
\sigma_i = \sqrt{\frac{1}{m} \sum_{j=1}^{m} (X_{ij} - \mu_i)^2}
\end{equation}
\begin{equation}
\label{eq:min}
\min_i = \min_{j=1}^{m} X_{ij}
\end{equation}
\begin{equation}
\label{eq:max}
\max_i = \max_{j=1}^{m} X_{ij}   
\end{equation}
Where \( X_{ij} \) is the value of the \( j \)-th feature for the \( i \)-th sample, and \( m \) is the number of features. To create new features based on domain knowledge by calculating specific ratios of the original features, we can assume that \( \mathbf{X}_{a, b} \) denote the element-wise ratio of columns \( a \) and \( b \) of \( \mathbf{X} \):
\begin{equation}
\label{eq:dom}
\text{Feature}_k = \frac{X[:, a_k]}{X[:, b_k]}    
\end{equation}
Table \ref{table:pairs} shows how we can choose a set of index pairs \( (a_k, b_k) \) based on domain knowledge.
\begin{table}
\caption{Index pairs for domain-specific features.}
\label{table:pairs}
\centering
\begin{tabular}{|c|c|}
\hline
$a_k$ & $b_k$ \\
\hline
3 & 5 \\
2 & 1 \\
5 & 2 \\
7 & 6 \\
3 & 6 \\
3 & 4 \\
0 & 5 \\
8 & 2 \\
7 & 4 \\
0 & 6 \\
4 & 1 \\
5 & 1 \\
8 & 3 \\
6 & 1 \\
\hline
\end{tabular}
\end{table}
We need to combine the original polynomial features, statistical features, and domain-specific features into a combined feature matrix \( \mathbf{X}_{\text{comb}} \):
\begin{equation}
\label{eq:combined}
\mathbf{X}_{\text{comb}} = \text{concat}(\mathbf{X}_{\text{poly}}, \mu, \sigma, \min, \max, \text{F}_1, \text{F}_2, \ldots, \text{F}_{14})   
\end{equation}
Where \( \text{concat} \) denotes the concatenation operation along the feature axis. \( \mathbf{X}_{\text{comb}} \) is the final expanded feature matrix including all derived features. \( \mathbf{X}_{\text{poly}} \) represents the polynomial features. \( \mu, \sigma, \min, \max \) are the statistical features (mean, standard deviation, minimum, and maximum). \( \text{F}_1, \text{F}_2, \ldots, \text{F}_{14} \) represents the domain-specific features generated as ratios of the original features \( a_k \) and \( b_k \) as described in Table \ref{table:pairs}. Algorithm \ref{algo} represents the pseudocode of the AquaAugmentor algorithm, Table \ref{table:complexity} analyzes the computational complexities, and Fig. \ref{fig} illustrates the workflow of the AquaAugmentor algorithm. 
\begin{algorithm}
\caption{AquaAugmentor Algorithm}
\label{algo}
\begin{algorithmic}[1]
\REQUIRE Dataset $D$ with $n$ samples and $m$ features, target variable $y$ in the last column
\ENSURE Combined feature matrix $X_{\text{comb}}$
\STATE \textbf{Step 1: Separate Features and Target}
\STATE $X = D[:, :m]$, $y = D[:, m]$
\STATE \textbf{Step 2: Create Polynomial Features}
\STATE $poly = \text{PolynomialFeatures}(\text{degree}=d,\newline \text{interaction\_only}= \text{True}, \text{include\_bias}= \text{False})$
\STATE $X_{\text{poly}} = poly.fit\_transform(X)$
\STATE \textbf{Step 3: Add Statistical Features}
\STATE Initialize $stat\_features$
\FOR{each sample $i$ in $X$}
    \STATE Calculate $\mu_i$, $\sigma_i$, $\min_i$, $\max_i$
    \STATE Append $\mu_i$, $\sigma_i$, $\min_i$, $\max_i$ to $stat\_features$
\ENDFOR
\STATE \textbf{Step 4: Add Domain-Specific Features}
\STATE $index\_pairs = \{(3, 5), (2, 1), (5, 2), (7, 6), (3, 6), (3, 4),\newline (0, 5), (8, 2), (7, 4), (0, 6), (4, 1), (5, 1), (8, 3), (6, 1)\}$
\STATE Initialize $domain\_features$
\FOR{each pair $(a_k, b_k)$ in $index\_pairs$}
    \STATE Calculate $feature_k = \frac{X[:, a_k]}{X[:, b_k]}$
    \STATE Append $feature_k$ to $domain\_features$
\ENDFOR
\STATE \textbf{Step 5: Combine All Features}
\STATE $X_{\text{comb}} = \text{concat}(X_{\text{poly}}, \mu, \sigma, \min, \max, \text{feature}_1, \newline\text{feature}_2, \ldots, \text{feature}_{14})$
\RETURN $X_{\text{comb}}$
\end{algorithmic}
\end{algorithm}
\begin{table}
\caption{Computational complexity of the AquaAugmentor algorithm.}
\label{table:complexity}
\centering
\begin{tabular}{|>{\raggedright}p{0.8cm}|>{\raggedright}p{2cm}|>{\centering}p{2cm}|>{\centering\arraybackslash}p{2cm}|}
\hline
\textbf{Step} & \textbf{Operation} & \textbf{Time Complexity} & \textbf{Space Complexity} \\
\hline
\textbf{1} & Splitting dataset into features and target & \( O(n) \) & \( O(nm) \) (features), \( O(n) \) (target) \\
\hline
\textbf{2} & Generating polynomial features & \( O(n \cdot m^d) \) & \( O(n \cdot m^d) \) \\
\hline
\textbf{3} & Calculating mean, std, min, max & \( O(nm) \) (each) & \( O(n) \) (each) \\
\hline
\textbf{4} & Creating domain-specific ratios & \( O(n \cdot k) \) & \( O(n \cdot k) \) \\
\hline
\textbf{5} & Concatenating all features & \( O(n \cdot (m^d + k + 4)) \) & \( O(n \cdot (m^d + k + 4)) \) \\
\hline
\textbf{Overall} & All steps combined & \( O(n \cdot m^d) \) & \( O(n \cdot m^d) \) \\
\hline
\end{tabular}
\end{table}
\subsection{Data Balancing and Normalization}
After applying the feature augmentation, the number of features increased from 9 to 62. We then combined the target attribute with the augmented features and applied the SMOTE to balance the class distribution. We split the dataset into training, validation, and test sets in a 70:15:15 ratio. Finally, we normalized the data using StandardScaler to ensure that each feature contributed equally to the model's performance.
\subsection{Predictive Modeling}
We employed a variety of machine learning and deep learning models to classify water potability. The machine learning models used include Linear Regression, Random Forest, Linear SVM, KNN, MLP, CatBoost, AdaBoost, XGBoost, GBM, Naive Bayes, and Logistic Regression. In addition, several deep learning models were implemented, such as 1-D CNN, LSTM, Autoencoder, Variational Autoencoder, TabNet, TabTransformer, and a hybrid CNN-LSTM model. These models were selected to evaluate their effectiveness in accurately classifying water quality and to determine the impact of our novel feature augmentation algorithm, AquaAugmentor, on their predictive performance.
\section{Results Analysis}
\label{sec:4}
\begin{figure}[H]
\centering
\includegraphics[width=\linewidth]{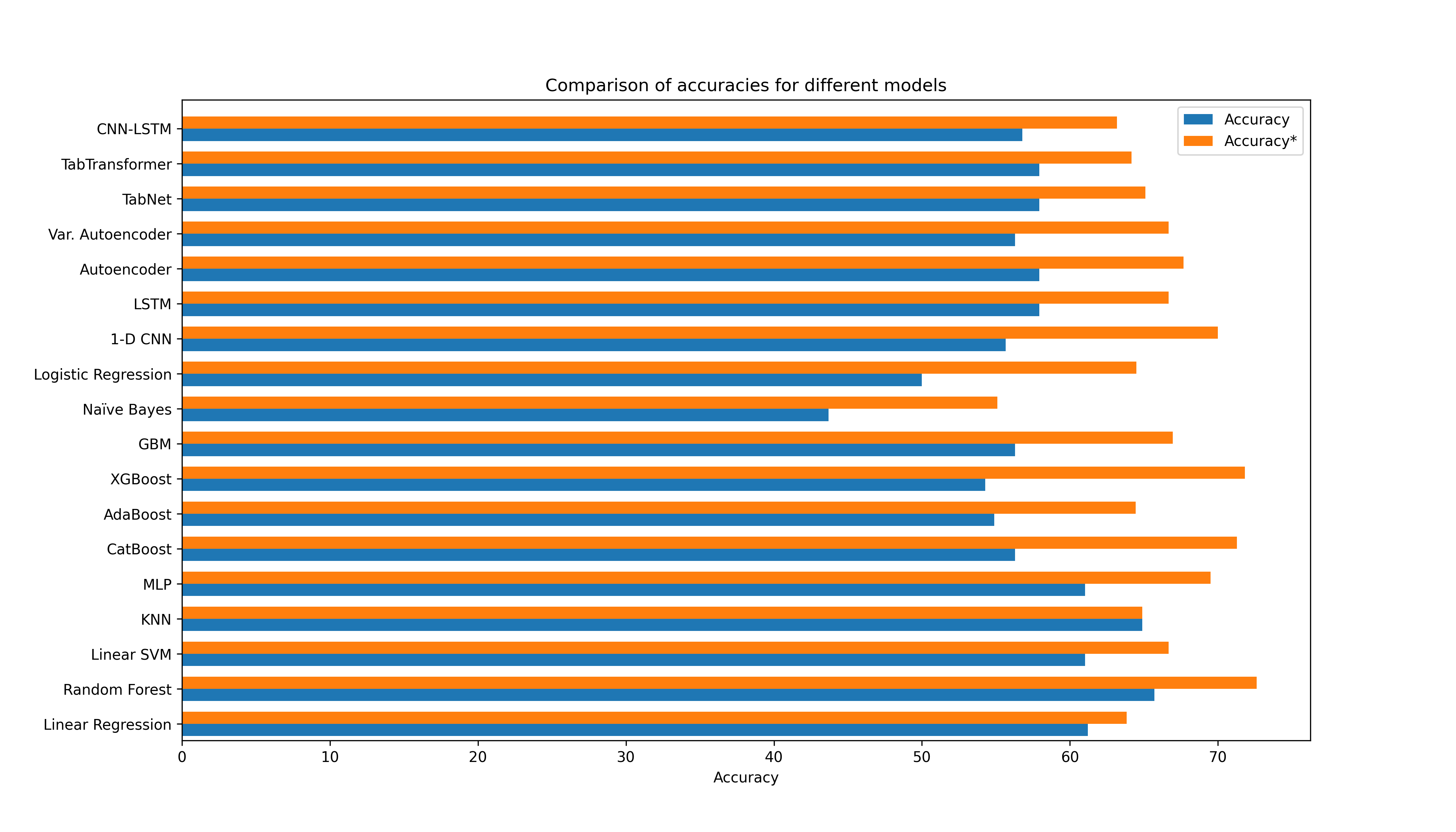}
\caption{Comparison of test accuracies for different models.}
\label{fig2}
\end{figure}

\begin{figure}
\centering
\includegraphics[width=\linewidth]{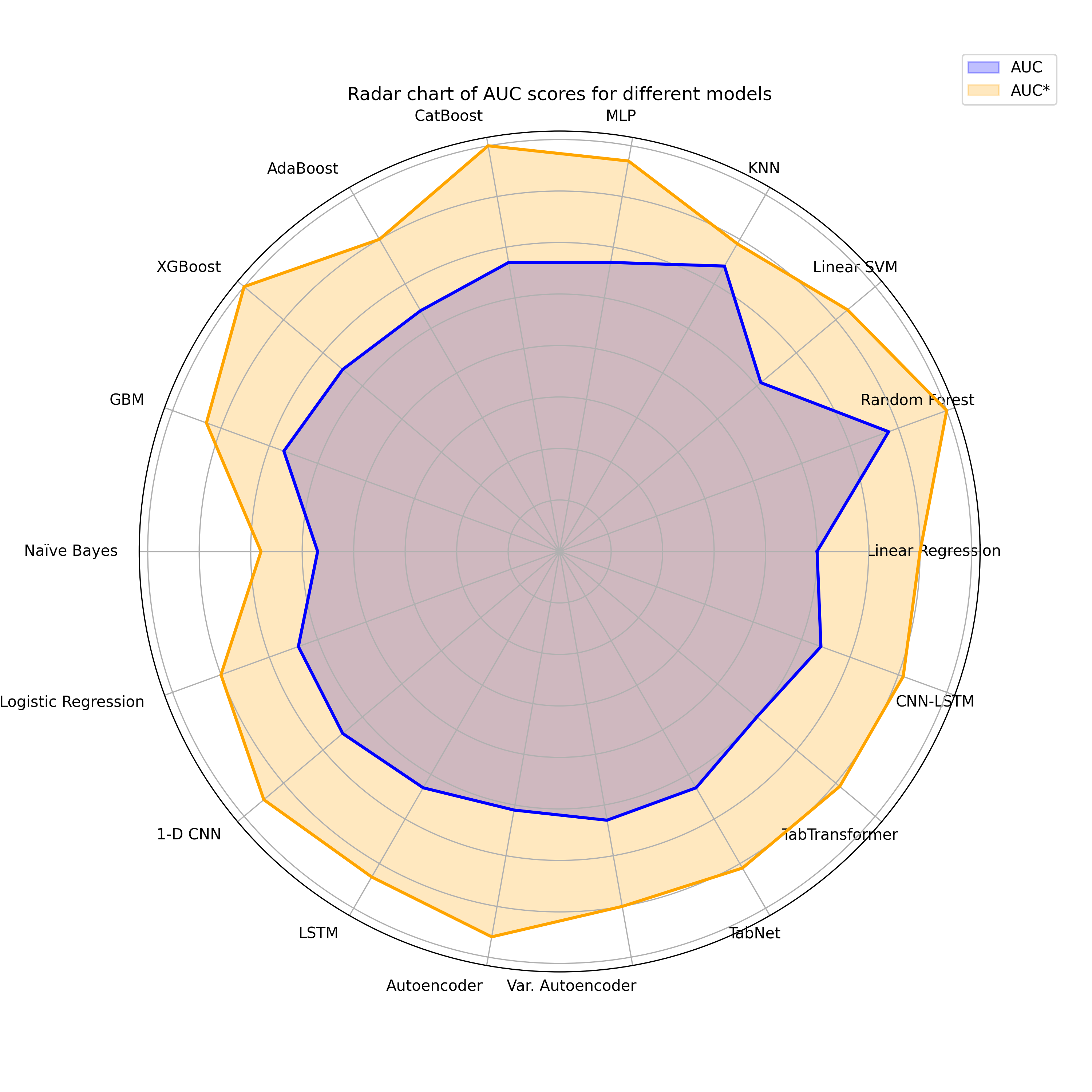}
\caption{Comparison of AUC scores for different models.}
\label{fig3}
\end{figure}

\begin{figure}
\centering
\includegraphics[width=\linewidth]{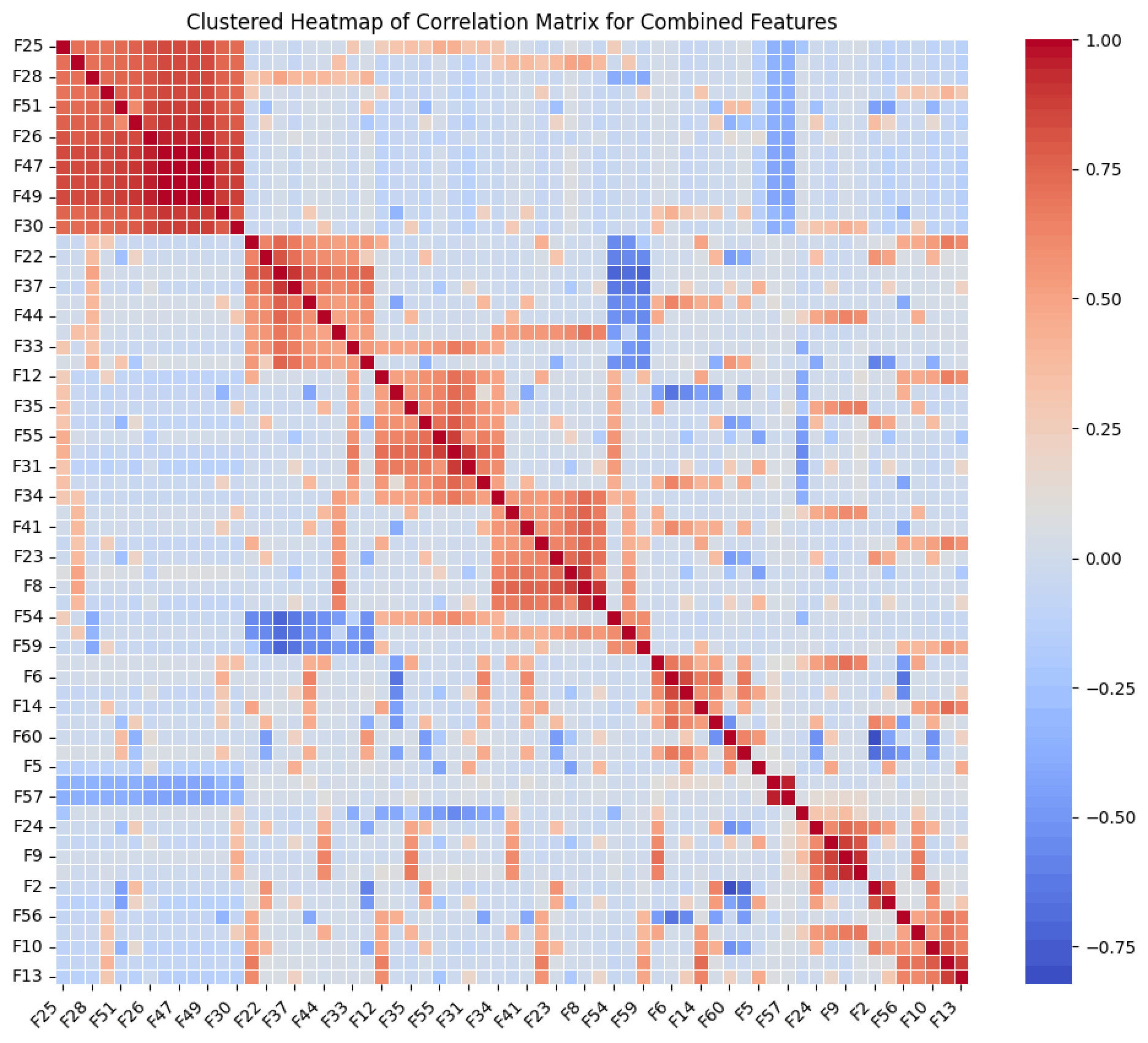}
\caption{Clustered correlation heatmap matrix after applying the AquaAugmentor algorithm.}
\label{fig4}
\end{figure}

\begin{figure}
\centering
\includegraphics[width=\linewidth]{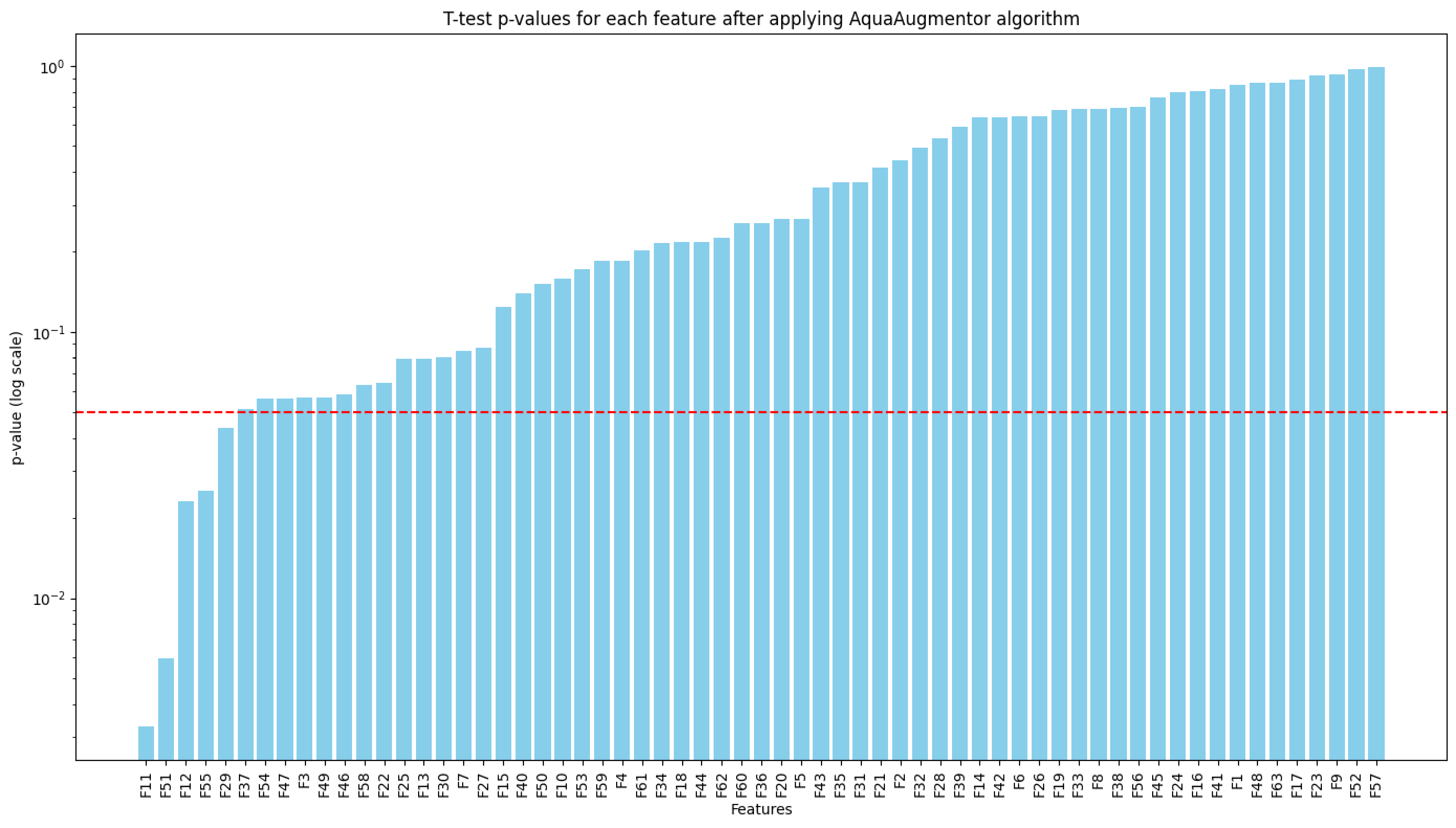}
\caption{t-Test p-values for each feature after applying the AquaAugmentor algorithm.}
\label{fig5}
\end{figure}

The results shown in Table \ref{results_table} and the accompanying graphs highlight the significant influence of the AquaAugmentor algorithm on the effectiveness of several machine learning and deep learning models. This approach improves accuracy and increases the AUC scores of most models. The accuracy of the Random Forest model is significantly enhanced from 65.71\% to 72.62\% and its AUC score from 0.68 to 0.80. This significantly enhances the model’s capacity to classify samples of drinkable and non-drinkable water accurately. Similarly, the XGBoost model also got a significant increase in accuracy, rising from 54.27\% to 71.37\%. Additionally, its AUC score improved from 0.55 to 0.80, indicating a notable enhancement in prediction capability and reliability due to using AquaAugmentor. Linear Regression, which typically faces challenges when dealing with complex datasets, also saw substantial advantages. The model's accuracy improved from 61.22\% to 63.83\%, while the AUC score significantly increased from 0.50 to 0.70, showing a substantial improvement in distinguishing between potable and non-potable classes. The accuracy of models such as LSTM and Autoencoder demonstrated significant enhancements. LSTM's accuracy increased from 57.93\% to 66.67\%, and its AUC improved from 0.53 to 0.71. The Autoencoder, a model proficient in dealing with intricate patterns, improved its accuracy from 57.93\% to 67.67\% and its AUC from 0.51 to 0.80, showcasing AquaAugmentor's capability to boost deep learning models.\newline
\begin{table}
\centering
\caption{Results for different machine learning and deep learning models (* = with AquaAugmentor algorithm).}
\label{results_table}
\begin{tabular}{|>{\raggedright\arraybackslash}m{1.64cm}|>{\raggedright\arraybackslash}m{1cm}|>{\raggedright\arraybackslash}m{0.6cm}|>{\raggedright\arraybackslash}m{1.06cm}|>{\raggedright\arraybackslash}m{0.6cm}|}
\hline
Model & Accuracy & AUC & Accuracy* & AUC* \\ \hline
Linear Regression & 61.22\% & 0.50 & 63.83\% & 0.70 \\ \hline
Random Forest & 65.71\% & 0.68 & 72.62\% & 0.80 \\ \hline
Linear SVM & 61.02\% & 0.51 & 66.67\% & 0.73 \\ \hline
KNN & 64.90\% & 0.64 & 64.90\% & 0.69 \\ \hline
MLP & 61.02\% & 0.57 & 69.50\% & 0.77 \\ \hline
CatBoost & 56.30\% & 0.57 & 71.29\% & 0.80 \\ \hline
AdaBoost & 54.88\% & 0.54 & 64.44\% & 0.70 \\ \hline
XGBoost & 54.27\% & 0.55 & 71.83\% & 0.80 \\ \hline
GBM & 56.30\% & 0.57 & 66.94\% & 0.73 \\ \hline
Naive Bayes & 43.70\% & 0.47 & 55.09\% & 0.58 \\ \hline
Logistic Regression & 50\% & 0.54 & 64.50\% & 0.70 \\ \hline
1-D CNN & 55.66\% & 0.55 & 70\% & 0.75 \\ \hline
LSTM & 57.93\% & 0.53 & 66.67\% & 0.73 \\ \hline
Autoencoder & 57.93\% & 0.51 & 67.67\% & 0.76 \\ \hline
Variational Autoencoder & 56.30\% & 0.53 & 66.67\% & 0.70 \\ \hline
TabNet & 57.93\% & 0.53 & 65.11\% & 0.71 \\ \hline
TabTransformer & 57.93\% & 0.50 & 64.17\% & 0.71 \\ \hline
CNN-LSTM & 56.78\% & 0.54 & 63.17\% & 0.71 \\ \hline
\end{tabular}
\end{table}
The bar chart depicted in Fig. \ref{fig2} clearly illustrates the enhancements in test accuracies among various models, both with and without the implementation of AquaAugmentor. Improved models with AquaAugmentor routinely achieve better results than models without it, highlighting the algorithm's ability to enhance model performance. For instance, the increase in accuracy of the Random Forest model is noticeable, as is the improvement in the XGBoost and LSTM models. This graphic offers a concise and easily understandable comparison of the efficacy of AquaAugmentor. Fig. \ref{fig3} illustrates a radar chart illustrating the AUC scores, providing a comprehensive perspective on the model's performance across multiple dimensions. This graphic showcases the extensive enhancements made to several models, with notable increases in the areas shown by the radar plot for models improved with AquaAugmentor. Models such as CatBoost, XGBoost, and MLP demonstrate significant improvements in their AUC scores, highlighting the efficacy of AquaAugmentor in enhancing the model's capacity to differentiate between classes. The clustered heatmap in Fig. \ref{fig4} illustrates the interconnections among the different characteristics following augmentation. This heatmap visually represents the correlations between different features, providing insights into the relationships and interdependencies among the features after augmentation. Specifically, certain characteristics have strong connections, creating separate groups essential for the enhanced performance observed in the models. Furthermore, the t-test findings illustrated in Fig. \ref{fig5} offer a statistical confirmation of the relevance of the feature enhancements implemented by AquaAugmentor. The bar chart of p-values illustrates that numerous attributes have p-values below the 0.05 significance level, signifying that the disparities in these variables between potable and non-potable water samples are statistically significant. The statistical significance confirms the improvements in model performance and verifies the efficiency of AquaAugmentor in enhancing features.\newline
It is crucial to acknowledge that several research studies of different domains claim very high accuracies of 90\% or higher, and these findings are frequently not reproducible in practical situations due to environmental data's intricate and fluctuating nature. Real-world datasets, such as those employed for water potability classification, often contain noise and display substantial fluctuation, rendering the achievement of high accuracies impractical. On the other hand, the enhancements accomplished by AquaAugmentor are practical and show tangible usefulness. AquaAugmentor greatly improves models' accuracy and AUC scores, increasing their robustness and reliability. This makes them more appropriate for real-world applications where maintaining safe drinking water is crucial. When determining which models are appropriate, machine learning models such as Random Forest, XGBoost, and GBM benefit from AquaAugmentor since they help them properly manage enriched and sophisticated feature sets. Deep learning models, such as LSTM and Autoencoders, provide significant enhancements, suggesting their appropriateness for intricate pattern identification and feature interactions. Nevertheless, it is important to take into account the restrictions. The process of feature augmentation using AquaAugmentor can result in higher computing complexity and longer training durations, especially when applied to deep learning models. Furthermore, the method's efficacy may differ based on the excellence and characteristics of the initial dataset, requiring meticulous preprocessing and validation to guarantee optimal performance. Although AquaAugmentor has a few limitations, it is a valuable tool for improving the performance of models used in water quality categorization and other environmental data applications.
\section{Conclusions}
In this study, AquaAugmentor algorithm represents remarkable progress in environmental data science, namely in the classification of water potability, a major concern in the public health sector. This originality resides in its capacity to enhance low-dimensional datasets by adding features obtained from polynomial expansions, statistical measurements, and domain-specific information. This greatly improves the predictive performance of machine learning and deep learning models. The results exhibit consistent performance improvements across multiple models, highlighting AquaAugmentor's capacity to enhance accuracy and dependability in water quality predictions. The statistical validation using t-tests validated the importance of these changes, which is consistent with the observed performance benefits and reinforces the practical usefulness of AquaAugmentor. AquaAugmentor's practical advancements showcase its potential for effectively ensuring safe drinking water, in contrast to the frequently unexplained high accuracies mentioned in many research works. Further research could investigate adaptive feature augmentation methods to dynamically modify feature complexity according to dataset properties, reducing computing costs and training durations of the AquaAugmentor algorithm. Incorporating AquaAugmentor with real-time data collecting systems, such as IoT sensor networks, could facilitate ongoing monitoring and prompt analysis of water quality. This would offer timely insights and interventions.
\label{sec:5}

\section*{Acknowledgment}
We would like to express our gratitude to the Center for Research Innovation and Transformation (CRIT) at Green University of Bangladesh for their generous financial support.

\bibliographystyle{IEEEtran}
\bibliography{References}

\end{document}